\documentclass{sigchi} 
\usepackage{times}
\usepackage{helvet}
\usepackage{courier}
\usepackage{graphicx}
\usepackage{epsfig}
\usepackage{epstopdf}
\usepackage{latexsym}

\toappear{ } 

\usepackage{graphics} 
\usepackage{times}    
\usepackage{url}      

\makeatletter
\def\url@leostyle{%
  \@ifundefined{selectfont}{\def\UrlFont{\sf}}{\def\UrlFont{\small\bf\ttfamily}}}
\makeatother
\def\pprw{8.5in}
\def\pprh{11in}

\let\oldbibliography\thebibliography
\renewcommand{\thebibliography}[1]{%
  \oldbibliography{#1}%
  \setlength{\itemsep}{0pt}%
}

\newcommand{\nop}[1]{{}}

\newtheorem{example}{Example}

\begin{document}

\title{Using the Crowd to Generate Content for Scenario-Based Serious-Games}


\numberofauthors{3}
\author{
  \alignauthor Sigal Sina\\
    \affaddr{Bar-Ilan University, Israel}\\
    \email{sinasi@macs.biu.ac.il}\\
   \alignauthor Sarit Kraus\\
    \affaddr{Bar-Ilan University, Israel}\\
    \email{sarit@cs.biu.ac.il}\\
  \alignauthor Avi Rosenfeld\\
    \affaddr{Jerusalem College of Technology, Israel}\\
    \email{rosenfa@jct.ac.il}\\
}


\maketitle

\begin{abstract}
In the last decade, scenario-based serious-games have become a main tool for learning new skills and capabilities.
An important factor in the development of such systems is the overhead in time, cost and human resources to manually create the content for these scenarios.
We focus on how to create content for scenarios in medical, military, commerce and gaming applications where maintaining the integrity and coherence of the content is integral for the system's success.
To do so, we present an automatic method for generating content about everyday activities through combining computer science techniques with the crowd.
We use the crowd in three basic ways: to capture a database of scenarios of everyday activities, to generate a database of likely replacements for specific events within that scenario, and to evaluate the resulting scenarios. 
We found that the generated scenarios were rated as reliable and consistent by the crowd when compared to the scenarios that were originally captured.
We also compared the generated scenarios to those created by traditional planning techniques.
We found that both methods were equally effective in generated reliable and consistent scenarios,
yet the main advantages of our approach is that the content we generate is more varied and much easier
to create.
We have begun integrating this approach within a scenario-based training application for novice investigators within the law enforcement departments to improve their questioning skills.
\end{abstract}

\vspace{-5pt}
\section{Introduction}
\label{section:Introduction}

Simulations and scenarios-based games, which constitute an important subset of serious-games, are an important tool for learning new skills and capabilities. Such systems are currently being used in a broad range of applications such as military, government, educational and health care
\cite{Games:susi+2007,Games:graafland+2012,VH:gandhe+2009,VH:lane+2010}.
One main factor in the development of such systems is the overhead in time, cost and human resources to manually create the textual content for these scenarios.
Specifically, while the discipline of Procedural Content Generation (PCG) for Games  has focused on automatic generation of artificial assets such as textures, models, terrain and game rules, the automatic tools for scenario and textual content generation or adaptation for games and education are much less common \cite{Games:hendrikx+2013}. 
In this paper, we provide an automatic method for generating textual content about everyday activities through combining computer science techniques with the crowd-- an approach that can be generally applied to education, government or health care scenarios.

This paper focuses on how one can easily generate textual content within a scenario that can provide narratives within a variety of domains. Our approach presents a novel solution which takes manually written scenarios of everyday activities 
and uses it to automatically create  a new revised, reliable and consistent scenario.  At the core of our approach is a database of daily, social scenarios and narratives.  We demonstrate how this database is created from Amazon Mechanical Turk (AMT) workers, an established method for data collection from crowd \cite{crowd:AmazonTurk+2010}. We are able to ensure the consistency of these scenarios by providing the AMT works with a semi-structured form. This facilitates the creation of a varied database of daily activities' scenarios, their descriptions (narratives) written in natural language along with key attributes and statistical information regarding possible content replacements to the everyday activities scenarios. 
Once this database is large enough, we can match the best possibility for a given scenario using the k-nearest neighbor algorithm given the constraint that the replacement must preserve the integrity of the modified data and then generate a personalized description for it using a `fill and adjust" approach. Using crowdsourcing offers several advantages:
first, it enables us to construct the activities' scenarios and narratives database rapidly and at a low cost.
Second, it gives us a large and diversified content of daily activities from the workers. Last, we also use crowdsourcing techniques to quickly evaluate and demonstrate the efficacy of our approach in an extensive user
study with another pool of workers.

While the approach that we present is general and can be used in many scenario-based games, we specifically focused on training scenarios. Specifically, our motivation for generating everyday content comes from a joint project  \texttt{\textbf{VirtualSuspect}} with the law enforcement training department.
The project's purpose is to train new law enforcement detectives of property felonies to efficiently extract  relevant information from an interviewee.
As our approach generates content with relatively low cost and maintenance, we can easily add new training cases, allowing investigators to have repeated practice sessions using different types of investigation techniques for different cases of property felonies.
In the \texttt{\textbf{VirtualSuspect}} project, the generated content can also be applicable as an alibi for a specific case, for example, consider a case where a robber broke into a private house on Sunday
night and stole a laptop, jewelry and some cash money.
The law enforcement investigator (the human trainee) can question a suspect, focusing on what he did on Sunday night.
The interviewee (our virtual agent), which can be innocent or not, needs a coherent scenario of his Sunday activities that is consistent with the facts that are known to the investigator and/or are common knowledge.
While this paper focuses only on describing the interviewee's scenario generation portion of this project,
it is important to stress that our proposed approach can also be useful for other applications with everyday content, especially in training scenarios such as training doctors to ask a patient the right questions or to help train candidates in a job interview application.

We found that the scenarios and their activities' details (narratives) we generated were
rated as being as believable and consistent as the original scenarios and the
original activities' details, and also compared them to activities' details
that were generated with the more costly planning technique.
In addition to the low cost of this technique, another  main
advantage of our approach is that the generated activities' details of the
revised scenarios and narratives are much more diversified than those with a traditional planning
technique. This advantage is an important key when implemented on
large amounts of data, as one needs as varied a set of replacements
as possible in order to keep the modified scenarios believable.

\vspace{-5pt}
\section{Related Work}
\label{section:related-work}

The development of serious games for training is a complex and time-consuming process \cite{Games:nadolski+2008}.
While early works considered a whole story creation without getting any
real-time input from the user \cite{SG:Meehan_1977,SG:Turner_1993}, the
later works focus on interactive narratives where the user is part
of the story and can affect the plotline \cite{SG:Cavazza+2002,SG:RiedlYoung_2010,SG:WareYoung_2011,SG:RiedlStern_2006}. 
Some works use also hybrid methods whereby a predefined plot is
created and an autonomic agent can later add to or adapt the plot in
real-time \cite{SG:NiehausLiRiedl_2010}. However,
these studies focus on the general plot of the story but not on the
story's details, which were almost exclusively  manually created by
content experts. A second important direction is the Procedural Content Generation (PCG) for Games discipline. However, while there has been significant in PCG generation for serious games in the past decade,
to date, tools for textual content generation are still lacking \cite{Games:hendrikx+2013}.
To address this need, we created a framework for generating coherent scenarios,
which include a sequence of activity events (activities) and
also the activities' descriptions (narratives).

The term scenario have several definitions  depending on the context it is used.
According to Hendrikx et al. \cite{Games:hendrikx+2013} ``Game scenarios describe, often transparently to the user, the way and order in which game events unfold. Two types of game scenarios can be distinguished, abstract and concrete. The abstract game scenarios describe how other objects inter-relate. The concrete game scenarios are explicitly presented in the game, for example as part of the game narrative.''.
In this paper, we refer to concrete scenario types.
As such, we use the term scenario to describe a sequence of activity events (activities) and their descriptions (narratives). 

As is the case with several recent works \cite{SG:HajarnisRiedl+2011,SG:NiehausLiRiedl_2010,SG:ZookRiedl+2012}, 
we implement a hybrid method which constitutes a  ``fill and
adjust" semi-automatic narrative generation method. However, unique
to our work, we leverage the crowd to create the textual content.
%
The first step of our
task is to identify the activities that should be modified. This
problem has been studied extensively in database systems
\cite{CS:BB04a,CS:Bertossi2006}, and we built upon previous work \cite{CS:BiskupWiese08}
that uses a maximal satisfiability solver (MaxSat)  in order to identify these
activities. Specifically, we used the
off-the-shelf ``akmaxsat'' solver \cite{CS:Kuegel_2010}, which
was one of the winners at ``MaxSat 2010'' for the ``Weighted Partial MaxSat'' category.
Once the portions of the data that must be modified were identified,
we found that using crowdsourcing allowed us to create a database of
alternatives quickly and at a low cost. In crowdsourcing platforms,
such as AMT or MicroTask, requesters typically break up a complex
task into a number of simpler sub-tasks to make them easily
applicable for layman workers. Previous work found that although
micro-task markets have great potential for rapidly collecting user
measurements at low costs, special care is needed to formulate tasks
in order to properly harness the capabilities of the crowd
\cite{crowd:kittur+2008}. In order to facilitate the successful
creation of the database needed for our application, we built
questionnaires with precise instructions for each task and with
proper compensation for the workers.

One traditional approach for activity-details generation is
planning-based systems
\cite{SG:Meehan_1977,SG:RiedlYoung_2010,SG:WareYoung_2011,SG:Cavazza+2002}.
The planning-based approach uses a causality-driven search to link a
series of primitive actions in order to achieve a goal or to perform
a task. For the domain of descriptive, every day activities,
hierarchical scripts can capture common ways to perform the
activity. Therefore, we implemented the planning-based generator
using a Hierarchical Task Network (HTN), which is one of the
best-known approaches for modeling expressive planning knowledge for
complex environments. We used the state-of-the-art SHOP2 planner
\cite{SG:Nau+2003}, a well known HTN planner, which has been
evaluated and integrated in many real world planning applications
and domains including: evacuation planning, 
evaluation of enemy threats and manufacturing processes.
In order to validate the significance of modified scenarios with the
activity-details we created, we also implemented a planning-based
activity-details generator. As we later report, this approach is
more costly to implement  and generates less varied scenarios than the crowdsourced approach  we  now describe.

\vspace{-5pt}
\section{System Overview}

\begin{figure}[ht]
\vspace{-5pt}
\centering
\includegraphics[width=1.0\linewidth]{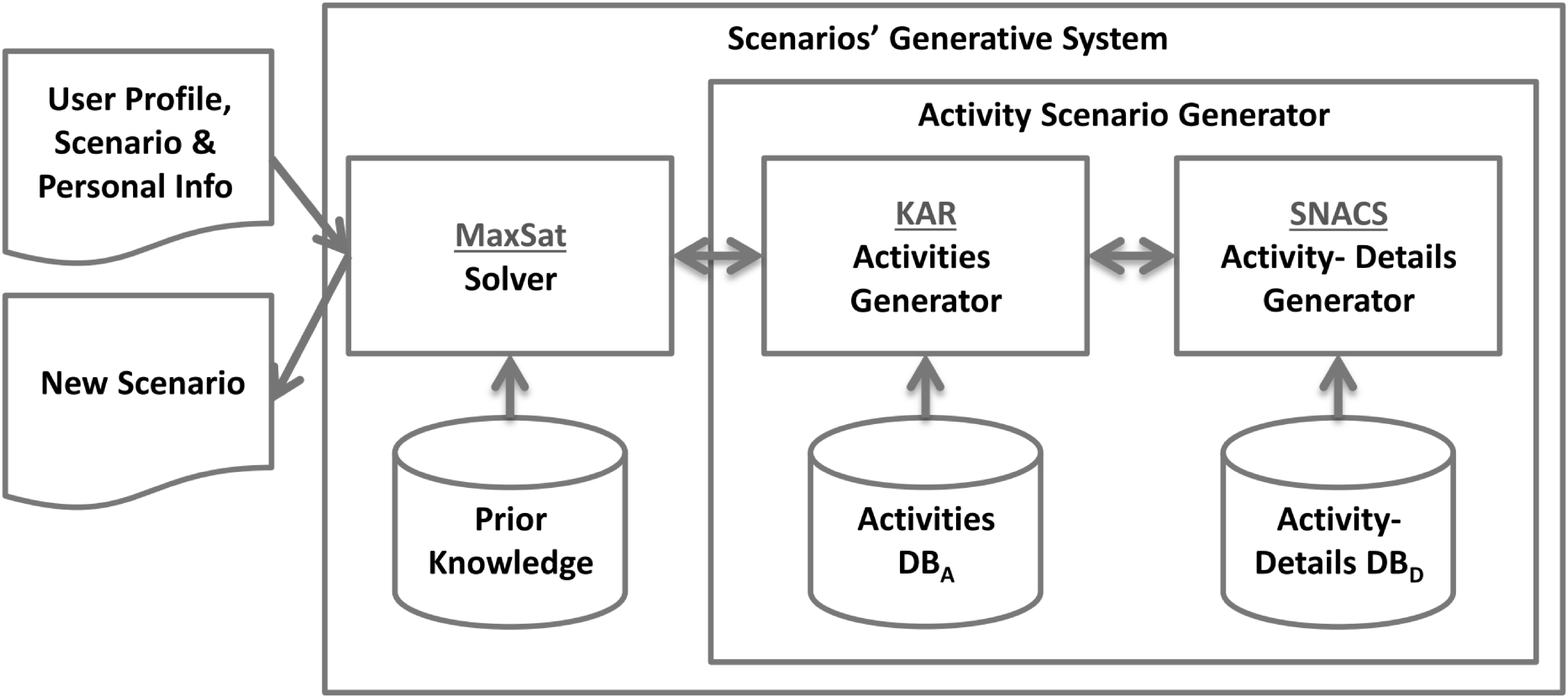}
\vspace{-15pt}
\caption{System Modules and Flow}
\vspace{-5pt}
\label{fig:sys-flow}
\end{figure}

We propose and build a system (presented in Figure
\ref{fig:sys-flow}) which ensures the scenario's integrity while
replacing a given set of information. In the
paper we will use the following running example to explain different
stages within the system. For clarity reasons, we are using a simple
example.
\vspace{-3pt}
\begin{example}
\label{examp:examp-1} John is a 21-year-old male who is single
and has no children. He broke into a private house on Sunday night
and stole a laptop, jewelry and cash money.
He is now being questioning and need an alibi story for Sunday night.
\end{example}
\vspace{-3pt}

In this example, our system needs to preserve John's
activity by concealing the information that John broke into a house.
This is done by replacing the activity $BrokeIntoHouse$ with a
common activity, such as $EatDinner$. After this activity switch
has been made, a modified
scenario with a new activity-details will be generated. Our
system does this by basing itself on details from a collection of
reported activities, which it modifies to better
match the scenario main character's profile. 
Referring back to example \ref{examp:examp-1}, we base the revised
scenario on a reported activity of a 26-year-old male with no children
who goes out to lunch (Example \ref{examp:event-exp1}).
Note that in this case we need to change the details about the time
and location to match the required alibi (Example \ref{examp:event-exp2}).
\vspace{-2pt}
\begin{example}
\label{examp:event-exp1} ``I went and got lunch and a beer at a local bar ``The Liffey".
It was during March Madness, so I was watching some basketball.
I sat at the bar and got chicken wings. I watched a few basketball games and ate.
I read the newspaper a bit too.
The food at ``The Liffey" is always good.
The team I picked won so that was also good.  "
\end{example}
\vspace{-2pt}
\vspace{-2pt}
\begin{example}
\label{examp:event-exp2}: ``On Sunday night I went out for dinner.
I did not really want to spend too much so I went to ``54th Street".
I sat at the bar and got chicken wings. I watched a few basketball games and ate.
I read the newspaper a bit too.
The food at ``54th Street" is always good.
The team I picked won so that was also good."
\end{example}
\vspace{-2pt}

We now focus on how
the system identifies what needs to be modified in the scenario, how
it finds a similar record on which to base the modified scenario, and how it
generates the final scenario with the appropriate activity-details.
As we analyzed the coherent problem in scenarios, we identified
three main questions the system needs to address. These questions
are: (1) What should be removed from the scenario? (2) With what
should we replace it? (3) How should we replace it? Accordingly, our
system consists of three main modules, one to address each of these
questions. The modules are: (1) A Maximal Satisfiability Solver
(MaxSat) which ensures the scenario's integrity and identifies the
places where modifications are required. Referring back to our
example, this module will identify the need to replace the
$BrokeIntoHouse$ activity; (2) Activities Generator which uses an
algorithm based on the k-nearest neighbor algorithm in order to choose the most appropriate activity replacement for the required modification. In our example, it suggests replacing the $BrokeIntoHouse$ activity with the
$EatDinner$ activity; and (3) Activity-Details Generator which expands
the required activity replacement into a realistic, reliable,
descriptive activity, as can be seen in examples  \ref{examp:event-exp1} and
\ref{examp:event-exp2}.

In this paper we do not focus on the MaxSat solver as this approach
has been previously studied \cite{CS:BiskupWiese08}.
We use an off-the-shelf MaxSat solver to generate the
optimal solution with the minimum number of modifications to the scenario.
We expand the traditional usage of this approach
by using placeholders to protect facts which will later be replaced.
Once the constraints have been identified, the system uses the
activities generator and activity-details generator modules to fill in
the missing details in the scenario for these placeholders.  After the scenario has
been finished but before the modified scenario is returned, the MaxSat
solver is run again in order to confirm that no constraints have been
violated in the final modified scenario.

This paper focuses on a novel approach using crowdsourcing
techniques to generate replacement activities and their details.
This is done through creating an activities dataset that contains a
collection of daily schedule records. These records are composed of
a list of activities and a collection of possible activities
replacements, which can form the skeleton of a modified scenario. The
details in these activities are selected from a dataset containing a
collection of descriptive reported activities written in natural
language.  The selected replacement activity is associated with
revised details that are consistent with a specific user profile. We
now detail how these modules are defined and applied.

\vspace{-5pt}
\section{Formal Definition of a Scenario}
\label{section:Formal-Definition}
Before we present the system implementation, we define several
concepts to be used in its description.
\begin{itemize} \parskip0pt \parsep0pt \topsep0pt \itemsep0pt
\item \texttt{User Profile} \texttt{(P)} - describes the user (i.e. the scenario's writer or the scenario's subject) properties and consists of gender, age, personal status and number of children. 
\item \texttt{Scenario} - a sequence of activities and their descriptions represented as a list of pairs \texttt{<AI, ADR>} where each activity instance \texttt{AI} is accompanied with an activity-details record \texttt{ADR}. The description of these two fields follows.
\item \texttt{Activity Instance} \texttt{(AI)} - is a specific occurrence of activity which is part of the scenario and is composed of the activity name, a day, start and end time, location and participants.
    In our example: $AI(night,$ $ John,$ $BrokeIntoHouse,$ $Downtown, alone)$.
\item \texttt{Activity-Details} \texttt{Record} \texttt{(ADR)} - is a tuple \texttt{<P,ADA,ADP>} where:
    \texttt{P}  is a user profile,
    \texttt{ADA} is the activity-details attributes vector
    and \texttt{ADP} is the activity natural language presentation.
    A detailed description  of the latter two fields follows immediately.
\item \texttt{Activity-Details} \texttt{Attributes} \texttt{(ADA)} -  contains a vector of attributes which
accompanies the activity-details.
This vector is a superset of the activity instance \texttt{AI}, which
contains the general attributes such as participants, a day and location, but it also contains information specific to the activity-details domain, such as restaurant name and type.
    It can contain optional values, and thus can be full or partial, for example in the eat-at-a-restaurant activity a person can eat at a restaurant alone, but can also go with a spouse.
    Within example \ref{examp:event-exp1} above, we represent this vector as:
    $\langle$day (Thursday),  part-of-day (noon), name (The Liffey),
    type (Bar and Grill), location (downtown)
    and participants (alone)$\rangle$.
\item \texttt{Activity} \texttt{Presentation} \texttt{(ADP)} - is the activity's detailed
    description written in natural language.
    It is composed of three parts:
    (1) The activity \texttt{Introduction} describes the main facts of the activity,
    such as who went, when, what are the main objects' names (which movie/restaurant), where and why;
    (2) The activity \texttt{Body} describes the activity in detail,
    what was the course of events and what happened during the activity;
    and (3) The activity \texttt{Perception}  describes how good or bad the experience
    was from the user's perspective.
    Note that we intentionally split the activity presentation into these three parts.
This semi-structured free text writing is very applicable when describing social, everyday
situations. 
It also centralizes most of the activity specific details in the
introduction part, which facilitates adjusting the activity to a new
user profile and attributes vector.
        Accordingly, the presentation of example \ref{examp:event-exp1} is:
        (1) \texttt{Introduction}: ``I went and got lunch and a beer at a local bar$\ldots$"
        (2) \texttt{Body}: ``I sat at the bar and got chicken wings$\ldots$"
        and (3) \texttt{Perception}: ``The food at ``The Liffey" is always good$\ldots$"
\end{itemize}

\section{System Implementation}
\label{section:system-impl}

We developed an innovative methodology to build the datasets using crowdsourcing. 
In all tasks, the AMT workers were first asked to provide their profiles \texttt{P} (gender, age,
personal status and number of children). Then, they were presented
with a semi-structured questionnaire containing a list of questions
and were asked to fill it out. Examples of these forms can be found at
http://aimamt.azurewebsites.net/.

One main challenge is how to best select
the most appropriate record from within the entire dataset.
To accomplish this task, we
define a compatibility-relevant measure (as we describe  in the
algorithm flow) which is based on the similarity measure between
attributes in order to predict which record is the best replacement.
The basic component of the similarity comparison is the decision whether
two values are similar. To make this comparison we associate each
one of the attributes with a specific comparison function which gets as input two values and returns one of the three
values: \textbf{same}, \textbf{similar} and \textbf{other}. 
Note that in the case one of the values is missing, it returns the \textbf{similar} value,
as we assume the generators will fill this attribute with a similar value.
For example, we consider the number-of-children attribute to be the \textbf{same} if the difference between the two values is 1 or less, \textbf{similar} if it is
less than or equal to 3 and \textbf{other} if one person has children and the other does not or when
the difference is greater than 3.
For the day attribute, the comparison function returns \textbf{same} if both values equal, \textbf{similar} if both values are weekdays or weekends and \textbf{other} otherwise.

\vspace{-5pt}
\subsection{Activities Generator}
\label{section:Activities-Generator} The activities generator's goal is to find the most appropriate activity replacement, such as $AI(night,John,$ $EatDinner,Downtown,alone)$, for the scenario's placeholder
provided by the solver, such as $AI(night,John,PH,$ $Downtown,alone)$. 
To accomplish this goal we built the KAR 
(\textbf{K}-nearest neighbor \textbf{A}ctivity \textbf{R}eplacement) generator whose input is a user's
profile, the scenario's activities list and an indication which activity
instance to needs be replaced. KAR returns a revised scenario with a replaced
activity which will later be associated with a natural language description and details.

\begin{figure*}[ht!]
\centering
\includegraphics[width=0.95\linewidth]{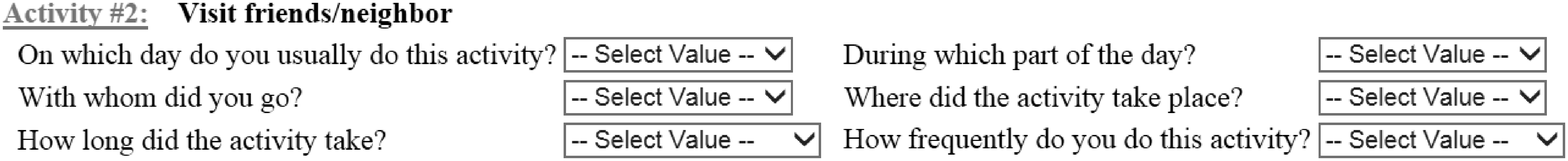}
\vspace{-5pt}
\caption{Activities Form}
\vspace{-10pt}
\label{fig:act-form}
\end{figure*}

\textbf{\texttt{The Dataset}} The activities dataset, denoted
$DS_A$, contains a collection of daily schedule records,
\texttt{SR}, and activity records, \texttt{AR}. The \texttt{SR} is a
tuple \texttt{<P,Sch>} where \texttt{P} is a user profile and
\texttt{Sch} is a daily schedule represented as a list of activity
instances \texttt{AI}. The \texttt{AR}  is a tuple \texttt{<P,Act>}
where \texttt{P} is a user profile and \texttt{Act} is the activity
properties and consists of the activity name (such as see-a-movie or
have-a-meeting) and six attributes: a day (a weekday or weekend),
part of day, duration, location, participants and frequency.
We use two types of questionnaires in order to acquire the two
record types. The first form is used to define the set of possible
activities and the second form is used to collect additional data on
each of the activities from a variety of profiles for the
activities generator (described below). In the first questionnaire,
we acquire weekday and weekend schedules (in an hour resolution),
where we asked the workers to describe the activities as
specifically as possible and limited each activity to up to a 3 hour
duration. As we later describe, this data was also used to evaluate
our system. For each activity in the schedule, the worker is asked
to fill in the activity name (written in free text), the participants
and the location of the activity.
Defining the set of possible activities requires only a few
schedules, and thus we collected 16 schedules from 8 subjects (4 male,
4 female, ages 23-53) which were paid 25-40 cents each for writing
two schedules. We then map all of the activities in these schedules
into an enumerated list and store the converted schedules
\texttt{Sch} with their profile \texttt{P} at $DS_A$ as the
\texttt{SR} records. The second questionnaire used to acquire
additional properties, \texttt{Act}, for each of the activities
in the collected schedules.
The workers were presented with a form which included a list of
activity record fields (Figure \ref{fig:act-form}), and were
asked to fill it out using predefined selection lists.
The \texttt{AR} records were collected from from 60 subjects (23 male, 37 female, ages 21-71) which were paid 35 cents each and were also stored at $DS_A$.

\textbf{\texttt{KAR}} Our activities generator, KAR, is implemented using the k-nearest neighbor algorithm
and it uses a compatibility measure to predict which activity record
is the best replacement for the placeholder. To calculate this
measure, we select 10 attributes: the 4 attributes of the profile
\texttt{P} and the 6 attributes of the activity \texttt{Act}.
KAR first calculates the similarity measure of each
of these attributes for all of the activity records \texttt{AR}
within the dataset $DS_A$ compared to the given user's profile
\texttt{P} and the activity placeholder \texttt{AI}, which it needs
to replace. For example \ref{examp:examp-1} with a profile (Male,
21, single, no children) and a placeholder
$AI(night,$$John,$$PH,$$Downtown,$$alone)$ compared to the following
activity record $\langle$(Female, 31, married, 2 children),
(EatAtRest, weekend, night, one hour, downtown, spouse, once a
month)$\rangle$, the similarity measures are: $\langle$gender (other),
age (other), number-of-children (other), personal-status (other),
day (same), part-of-day (same), duration (similar), location (same),
participants (other), frequency (similar)$\rangle$.
The importance for any two values to be the same or at least similar
depends on the specific attribute. For example, having the same
gender value in the generated scenario is much more important than
having the same age. To associate different importance levels for
each attribute similarity measure, we developed a scoring function
that gets an attribute and a similarity measure and returns a score
within the range [-15,15]. We refine this score function using
several preliminary trial and error iterations. The KAR generator then
calculates the compatibility measure for each of the records,
\texttt{AR}, as a summation of the scores of each of
these attributes and its calculated similarity measure. Last, KAR
sorts the activities records according to this measure and uses the
k-nearest neighbor algorithm in order to choose the best candidate.
Specifically, we implemented two variations of this algorithm, one
with K=1 and the other with K=11. While the K=1 variation returns
the activity with the highest measure, the K=11 variation also takes
into account the number of similar records and thus returns the
activity with the highest probability from the top 11 measures.

\vspace{-5pt}
\subsection{Activity-Details Generator}
\label{section:Alg-Details-Generator} The activity-details generator
module is responsible for turning a given activity instance in the revised
scenario into a realistic, reliable, descriptive activity.
It gets as input the user's profile \texttt{P} and a partial activity
details attributes vector \texttt{ADA} (which is made up of the values
given in the activity instance \texttt{AI}).
It returns as output a new activity-details record \texttt{ADR} which
contains a reasonable, consistent and realistic activity presentation
\texttt{ADP} written in natural language,
which substitutes activity in the revised scenario.
We implemented two types of generators: our approach, SNACS
(\textbf{S}ocial \textbf{N}arrative \textbf{A}daptation using \textbf{C}rowd\textbf{S}ourcing),
which uses the activity-details records we collected from the crowd, and for comparison
a traditional planning-based generator, which is a common technique for
content generation in many real world domains.

\textbf{\texttt{The Dataset}} We again use the crowd as the
source of the activity-details dataset, denoted as $DS_D$, and
build a collection of human activity-details written in natural
language for a specific activity, such as see-a-movie. We used a
dedicated, semi-structured questionnaire on AMT to collect the
activity-details record \texttt{ADR} which includes: the profile
\texttt{P}, the activity attributes vector \texttt{ADA} and the
activity presentation in natural language \texttt{ADP}. Here, the
workers were asked to describe daily, social activities in natural
language in as much detail as possible according to the three
activity presentation parts - introduction, body and perception.
Then, the workers were presented with a list of specific questions
used to collect the activity-details attributes vector, such as
``What was the name of the movie/restaurant?",  ``With whom did you
go?"  and ``on what day?". The completed records \texttt{ADR} were
then stored at $DS_D$. We intentionally split the activity's
detailed description into three parts. On one hand, this
semi-structured free text writing is very applicable  when
describing social, everyday situations, and it helps us to elicit a
detailed description of the activity from the workers. On the other
hand, it centralizes most of the activity-specific details in the
introduction, which allows us to adjust the activity-details to
a new user's profile and attributes vector during the activity
details generation without the need for intensive usage of NLP
tools. Specifically, we collect and store 10 activity-details for
4 activities: two are entertainment activities (see-a-movie and
eat-at-a-restaurant) and two are errand activities (buying-groceries
and dry-cleaning). These records were collected from 20 subjects (6
male, 14 female, ages 19-55) which were paid 50 cents each for
writing two activity-details.

\textbf{\texttt{SNACS}} Our activity-details algorithm, SNACS, first selects a candidate record
from the activity-details dataset $DS_D$.
We present 3 variations of this selection process below.
Then, the algorithm completes the missing activity-details attributes.
It generates attributes 
which are similar to the selected, original record's attributes and
matches them to the new user's profile. It starts with the
participant: who went and how many people participated in the
activity. It then generates the objects' names (movie, restaurant,
location) and time frame attributes. For example, if in the original
activity someone went to see a children's movie with his son and the
new user has no children, SNACS can choose to include his
niece/nephew among the participants. Next, the algorithm generates
the activity's natural language presentation. First, it replaces the
original activity's introduction, i.e. its first part (who went,
when, where, why), with a newly generated introduction according to
the new profile and the new vector of attributes. This is done by
using SimpleNLP \cite{NLG:GattReiter_2009}, a Natural Language
Generation (NLG), template-based surface realization, which creates
an actual text in natural language from a syntactic representation.
We created several NLG templates for each activity type, which were
randomly chosen during the introduction generation. For example, one
of the NLG templates used in order to build an activity introduction for
the see-a-movie activity was: ``Last $\left\langle time
\right\rangle$ I went to a movie with my $\left\langle with
\right\rangle$. We went to see the movie  $\left\langle movie
\right\rangle$ at $\left\langle theater \right\rangle$". Each such
template can generate a few variations according to the chosen
attributes. For example, the first part of the above template, where
the participants are a wife and son and the time is Sunday
afternoon, can generate (a) Last weekend I went to a movie with my
family or (b) Last Sunday afternoon I went to a movie with my wife
and my son. Finally, SNACS applies some adjustments to the body and
perception parts of the chosen activity's presentation (the second
and third parts). This is done by replacing the references of the
original attributes' vector with the new corresponding activity
attributes' vector. In example \ref{examp:event-exp2}, we replaced
the restaurant's name.

We implemented 3 variations of the SNACS algorithm which differ
in how the original candidate activity-details record is chosen:
\texttt{SNACS-Any}, \texttt{SNACS-Bst} and \texttt{SNACS-Tag}. The
\texttt{SNACS-Any} variation is a baseline measure that randomly
chooses one activity-details record from $DS_D$. No further logic is
performed to check how appropriate that choice is. In contrast, both
the \texttt{SNACS-Bst} and \texttt{SNACS-Tag} variations use a
compatibility measure to select which candidate from among all
records in $DS_D$ will serve as the base for the generated activity
description. The compatibility measure is based on 7 attributes: the
4 attributes of the profile \texttt{P} and only 3 attributes from
the activity-details attributes vector \texttt{ADA} (participants,
type and part-of-day). However, when assessing the compatibility of
the activity-details record \texttt{ADR}, we also have to account
for the activity presentation as written in natural language. Thus,
we define an importance level vector \texttt{ILV}, which corresponds
to these 7 attributes, for each activity-details record \texttt{ADR}
within the dataset \texttt{$DS_D$}. Each value in \texttt{ILV} is a
value \texttt{SM} and is used to represent the importance of the
compatibility of a given attribute within the activity body and
perception parts of the activity presentation. These values control
how much importance should be given to having similarity between the
original and generated activities' attributes. Accordingly, if a
given attribute within \texttt{ADR} can be modified without violating
any common sense implications, then the value is \textbf{other}. At
the other extreme, if that attribute is critical and even small
variations can make the activity implausible, then the value is
\textbf{same}. SNACS considers two approaches in which the vector
\texttt{ILV} can be built for every record. The first approach,
denoted as  \texttt{SNACS-Bst}, uses a fixed (automatic)
\texttt{ILV} across all records within \texttt{$DS_D$}.
Specifically, it contains the \textbf{same} value for the gender
attribute and a \textbf{similar} value for all of the other
attributes. The second approach, denoted as \texttt{SNACS-Tag},
utilizes a content expert to manually tag every record within
\texttt{$DS_D$}. For example, the manual \texttt{ILV} for example
\ref{examp:event-exp1} is $\langle$gender (same), age (similar),
number-of-children (other), personal-status (other), participants
(other), type (similar), part-of-day (similar)$\rangle$.

During runtime, SNACS first calculates the similarity measure of
each of these attributes for all of the
records \texttt{ADR} within the dataset $DS_D$ compared to the given
user's profile \texttt{P} and the (partial) activity instance
\texttt{AI} it needs to replace in order to select the best
candidate activity-details record. Recall that we used a
\textbf{similar} value in case of missing values. In example
\ref{examp:examp-1}, we evaluate John's profile and the activity
instance $AI(night,John,EatDinner,$ $Downtown,alone)$ compared to
the record \texttt{ADR} from example \ref{examp:event-exp1}. Thus, the
similarity measures are: $\langle$gender (same), age (similar),
number-of-children (same), personal-status (same), participants
(same), type (similar), part-of-day (other)$\rangle$. We again
build a score function (which was refined using trial and error iterations),
but this time it gets as input an attribute,
a similarity measure and an importance level and returns a score
within the range [-15,15].
SNACS then calculates the similarity compatibility measure for each
of the records, \texttt{ADR}, as a summation of the scores of each
of these attributes, its calculated similarity measure and its given
(fixed or manual) importance level. 
Finally, the record with the highest measure value is chosen as the best
activity-details candidate.

\textbf{\texttt{Planning-Based Generator}} In order to validate the
significance of SNACS, we also implemented a HTN planning-based
generator, denoted as \texttt{Planner}, using a
plot graph that we built manually. A plot graph \cite{SG:Weyhrauch_1997}
is a script-like structure, a partial ordering of basic actions that
defines a space of possible action sequences that can unfold during
a given situation.
As we wanted to get richer activity-details which include a detailed
description of the activity written in natural
language, we gave the planner an option to tailor natural language
descriptions in the basic actions portion of the activity.
We defined a set number, 10-15, of different descriptions that were
tailored to each one of the selected actions,
which assured the implemented planner had a variety of descriptions
with which to build activity-details.
These descriptions were manually handwritten by two experts, which
needed approximately one hour 
to write the set of descriptions for each basic action option. Part
of these descriptions were also manually tagged with specific tags,
such as movie or restaurant types. The tagging gave the generator an
option to choose between a generic description which can be
associated with any movie/restaurant type or a specific description
which can be associated with the current selected type, such as
action movie or Italian restaurant. Note that in SNACS, this step is
not necessary as it automatically gets the activity's detailed
descriptions from the original activity-details record. We also
implemented dedicated actions' realizators (SimpleNLP based) that
took the planner output, a semi-structured plan, and translated it
into a natural language activity presentation in addition to the
introduction's realizator we also used in SNACS.

Overall, the HTN-based generator has an inherently higher cost
associated with it as compared to SNACS for the following reasons: Both SNACS
and the planner have the steps of building the activity
introduction templates and the implementation of the logical
constraints. However, the planning-based algorithm implementation
also required the following additional manual steps: the manual
building of the plot graph; the writing, associating and tagging of
several detailed descriptions for each basic action; and writing a
specific realizator for each basic action. Each one of these steps
requires both time and resources from a content expert or a system's
designer. In fact, because of this cost overhead, we only used the
HTN-planner in order to define the two entertainment activities
(see-a-movie or eat-at-a-restaurant) and intentionally did not
implement the HTN-based generator for the errand activities.
Nonetheless, the activities' detailed descriptions produced by SNACS
were as good as those developed by this costly process, as our
results detail in the next section.

\textbf{\texttt{Random Baselines }}We also implemented two random methods as baselines to
ensure the validity of the experiment. The first random method,
denoted as \texttt{Rnd-SNACS}, uses the SNCAS infrastructure.
It randomly chooses one of the activity-details records in the dataset and
then randomly fills in the rest of the activity attributes.
The second random method, denoted as \texttt{Rnd-Planner}, uses the planning-based generator.
We removed the plan's logical built-in constraints and used random selections instead.

\vspace{-7pt}
\section{Experimental Evaluation}
\label{section:Experimental-Evaluation} In this section we present
the evaluation of the generated, revised scenarios and their associated descriptive activities.
Note that we don't include the evaluation of the MaxSat solver, as
we use an off-the-shelf, previously studied solver. We evaluated
separately the activities generator (KAR) and activity-details generator (SNACS) 
as described in the following sections. 
We use crowdsourcing to evaluate the efficacy of our
approach in an extensive user study (200 subjects), again in
AMT but with another pool of workers.
We ensure that subjects answer truthfully by including open test
questions and reviewing it manually before accepting the grades. We
also estimate that completing a survey should take 8-15 minutes, so
we filtered out forms which were filled out within less than 4
minutes.
Examples of these questionnaires can be found at http://aimamt.azurewebsites.net/.

\vspace{-8pt}
\subsection{Activities Generation Evaluation}
\label{section:Activities_Evaluation}
The purpose of this experiment is
to check the integrity and coherence of the scenario's activities
list after the replacement of one of its activities. 

\textbf{\texttt{Setup}} We chose to use the daily
schedules we already collected from the crowd for  $DS_A$, as the
original scenario (without the activity-details that will be evaluated
next).
We randomly cut a section of 7-8 hours from each of the
original activities list, to which we refer as \texttt{Original}. We
then randomly chose one activity for the list to be
replaced. We generated three revised lists by replacing the chosen
activity. Two of the lists were generated using KAR 
algorithm, one with K=1 and the other with K=11, to which
we refer as \texttt{KAR} \texttt{K=1} and \texttt{KAR} \texttt{K=11}
respectively. The third list \texttt{Random} was generated using a random
replacement and was implemented as a baseline to
ensure the validity the experiment. We then evaluated these 4
variations of each activities list using AMT questionnaires. Each
activities list was associated with a user profile and a day and the
workers were asked to rate it with reference to the profile
attached. The grades were valued from 1 (Least) to 6 (Most). An even
number of choices was given as we didn't want the users to choose a
middle value. The users were asked to grade three aspects:
reasonable, matching to profile and coherent.
The users were also asked
to try to recognize which, if any, activity was replaced, and
explain their answers in free text. Each activities list was rated
by 15-20 subjects, to ensure we had enough independent grades for
each generation method. A total of 80 subjects (38 male, 42 female,
ages 19-61) participated in the evaluation and were paid 25-40 cents
each.
\vspace{-8pt}
\begin{table}[htb]
\centering
\includegraphics[width=0.85\linewidth]{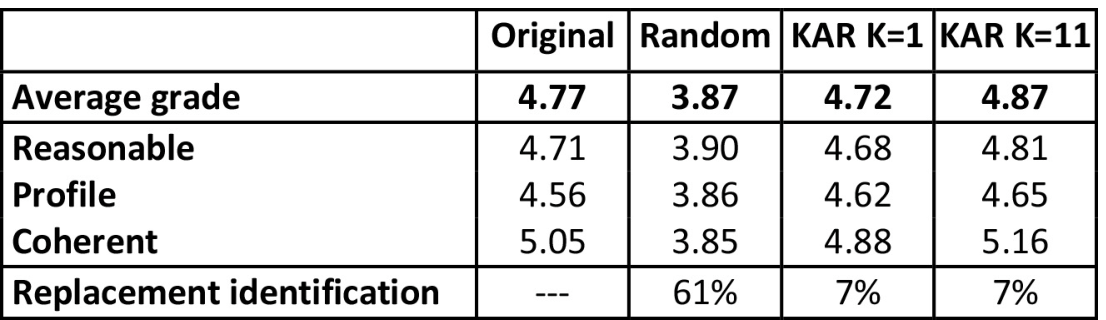}
\caption{Scenario�s Activities List Results}
\label{table:activities-grades-1}
\vspace{-10pt}
\end{table}

\textbf{\texttt{Results}} Table \ref{table:activities-grades-1}
presents the average grades for each
aspect and also the total average grade, which was calculated as the
average of these three grades. The results show that our generated
method  \texttt{KAR} \texttt{K=11} got the highest results, higher than
the \texttt{Original}, however, there is no significant difference
between the results of \texttt{Original}, \texttt{KAR} \texttt{K=11} and
\texttt{KAR} \texttt{K=1}.
As expected the \texttt{Random}
variation got lower grades, which are significantly lower than the
others (specifically, the ANOVA test of \texttt{Random} compared to
\texttt{Original}, \texttt{KAR} \texttt{K=11} and \texttt{KAR} \texttt{K=1}
had a much smaller than 0.05 threshold level with p=1.72E-9, 2.8E-11
and 1.13E-8 respectively).
It also can be seen from the replacement
identification percentage (the last row in Table
\ref{table:activities-grades-1}), that only 7\% of the users
identify the generated activity in the \texttt{KAR} \texttt{K=11} and
\texttt{KAR} \texttt{K=1} methods compared to 61\% in the \texttt{Random}
replacement. These percentages are significantly different from an
uniform random selection, which has a probability of 20\% for being
chosen.

\vspace{-8pt}
\subsection{Activity-Details Generation Evaluation}
\label{section:Details-Evaluation}
The purpose of this
experiment is to check the authenticity, integrity and coherence of
the generated descriptive activities which accompany the revised scenario.

\textbf{\texttt{Setup}} We chose to evaluate four
types of activities: see-a-movie, eat-at-a-restaurant,
buying-groceries and dry-cleaning.
For each activity type, we generated 4 user profiles.
Then for each profile we generated activity-details for all of the generation methods\footnote{We 
implemented the planning-based methods only for the see-a-movie and eat-at-a-restaurant activities because of the cost overhead.}.
We also randomly selected 4 additional activity-details out of the original activity-details dataset for each activity type.
As before, we ran the experiment using AMT questionnaires for comparison.
Each activity-details was associated with its user profile and the AMT workers were asked to rate (with the same 6-value scale as before) six aspects of the activity-details:
authenticity, reasonable, matching to profile, coherent, fluency and grammar.
As before, subjects were also asked to explain their choice in free text.
Each activity-details was rated by 8-10 workers to ensure we had enough independent grades for each activity type and generation method.
A total of 120 subjects (59 male, 61 female, ages 18-69) participated in the evaluation and were paid 40-50 cents each.
\vspace{-5pt}
\begin{table}[hb]
\centering
\includegraphics[width=0.95\linewidth]{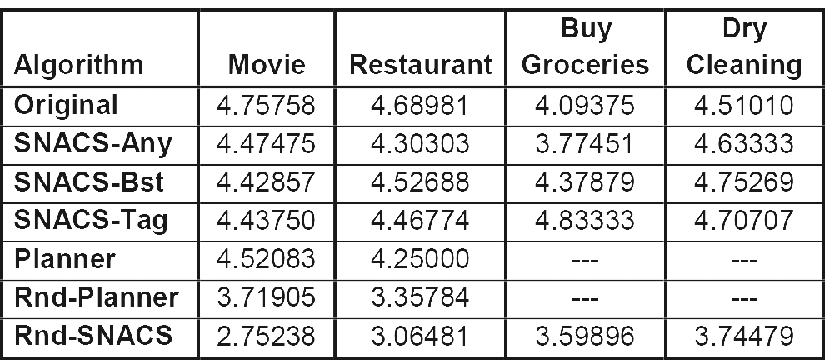}
\caption{Activity-Details (Narratives) Average Grades}
\label{table:average-grades-1}
\vspace{-10pt}
\end{table}

\textbf{\texttt{Results}} Table \ref{table:average-grades-1}
presents the average grade for each activity type and generation
method, which was calculated as the average of the six aspects'
grades. The results show that our approach produced revised
activity-details which were rated as being as reliable and consistent by
workers when compared to the original activity-details and the
planning-based technique. The main advantage of our approach is that
the descriptive activities are much easier to create. Both of the
random variations \texttt{Rnd-SNACS} and \texttt{Rnd-Planner} got
significantly lower grades than all the others (specifically, the
ANOVA test for the movie activity-details of \texttt{Rnd-Planner}
compared to \texttt{Original}, \texttt{Planner} and
\texttt{SNACS-Bst} had a much smaller than 0.05 threshold level with
p=1.95E-4, 4.99E-3 and 9.30E-3 respectively). The results also show
that all SNACS-based algorithms got very similar grades where
\texttt{SNACS-Bst} and \texttt{SNACS-Tag} are slightly better than
\texttt{SNACS-Any}. Overall for the entertainment activities
details, although the \texttt{Original} grades are slightly higher
than all of the other methods, there is no significant difference
between all of the SNACS-based generator methods or the
planning-based generator or the original activity-details. For the
errands activity-details, there is also no significant difference
between the grades of \texttt{SNACS-Tag}, \texttt{SNACS-Bst} and the
\texttt{Original} activity-details, although the
\texttt{SNACS-Any} variation got lower grades for the buy groceries
activity-details. We found that the results for each of the
six aspect grades were very similar to the average grade\footnote{We
omit this table due to lack of space.}.

\vspace{-8pt}
\section{Conclusions}
\label{section:Conclusions} The paper makes the following key two
contributions: (i) It is the first work to address the problem of
modifying scenarios to generate personal information but yet
maintains consistency even when varied scenarios are generated. (ii)
It provides a methodology to use crowdsourcing in a principled way
for this task. Instead of manually modifying scenarios, which makes
the development process costly in both time and resources, we used
the crowd as the source of our dataset, thus reducing the time and
effort needed by to maintain coherence in these scenarios.
To accomplish this task, we use the MaxSat logical engine in
combination with a novel approach for the generation of everyday
activities scenarios using the crowd. Our evaluation showed that our
revised scenarios and their activities' details were
rated as being as reliable and consistent as the original scenarios and the
original activities' details, and also compared them to activities' details
that were generated with the more costly planning technique.


\vspace{-8pt}
\section*{Acknowledgements}
This work was supported in part by ERC grant \# 267523.
%

\vspace{-8pt}
\bibliography{VirtualTrainer}

\begin{thebibliography}{10}

\bibitem{CS:Bertossi2006}
Bertossi, L.
\newblock Consistent query answering in databases.
\newblock {\em SIGMOD Record 35}, 2 (2006), 68--76.

\bibitem{CS:BB04a}
Biskup, J., and Bonatti, P.
\newblock Controlled query evaluation for enforcing confidentiality in complete
  information systems.
\newblock {\em International Journal of Information Security 3}, 1 (2004),
  14--27.

\bibitem{CS:BiskupWiese08}
Biskup, J., and Wiese, L.
\newblock Preprocessing for controlled query evaluation with availability
  policy.
\newblock {\em Journal of Computer Security 16}, 4 (2008), 477--494.

\bibitem{SG:Cavazza+2002}
Cavazza, M., Charles, F., and Mead, S.
\newblock Character-based interactive storytelling.
\newblock {\em IEEE Intelligent Systems 17}, 4 (2002), 17--24.

\bibitem{VH:gandhe+2009}
Gandhe, S., Whitman, N., Traum, D., and Artstein, R.
\newblock An integrated authoring tool for tactical questioning dialogue
  systems.
\newblock In {\em 6th Workshop on Knowledge and Reasoning in Practical Dialogue
  Systems} (2009).

\bibitem{NLG:GattReiter_2009}
Gatt, A., and Reiter, E.
\newblock {SimpleNLG: a realisation engine for practical applications}.
\newblock In {\em 12th European Workshop on Natural Language Generation}
  (2009), 90--93.

\bibitem{Games:graafland+2012}
Graafland, M., Schraagen, J., and Schijven, M.
\newblock Systematic review of serious games for medical education and surgical
  skills training.
\newblock {\em British Journal of Surgery 99}, 10 (2012), 1322--1330.

\bibitem{SG:HajarnisRiedl+2011}
Hajarnis, S., Leber, C., Ai, H., Riedl, M., and Ram, A.
\newblock A case base planning approach for dialogue generation in digital
  movie design.
\newblock In {\em Case-Based Reasoning Research and Development} (2011),
  452--466.

\bibitem{Games:hendrikx+2013}
Hendrikx, M., Meijer, S., Van Der~Velden, J., and Iosup, A.
\newblock Procedural content generation for games: a survey.
\newblock {\em ACM Trans on Multimedia Computing, Communications, and
  Applications\/} (2013).

\bibitem{crowd:kittur+2008}
Kittur, A., Chi, E.~H., and Suh, B.
\newblock Crowdsourcing user studies with mechanical turk.
\newblock In {\em SIGCHI Conference on Human Factors in Computing Systems}, ACM
  (2008).

\bibitem{CS:Kuegel_2010}
Kuegel, A.
\newblock Improved exact solver for the weighted max-sat problem.
\newblock {\em Workshop Pragmatics of SAT\/} (2010).

\bibitem{VH:lane+2010}
Lane, H.~C., Schneider, M., Michael, S.~W., Albrechtsen, J.~S., and Meissner,
  C.~A.
\newblock Virtual humans with secrets: Learning to detect verbal cues to
  deception.
\newblock In {\em Intelligent Tutoring Systems}, Springer (2010), 144--154.

\bibitem{SG:Meehan_1977}
Meehan, J.
\newblock Tale-spin, an interactive program that writes stories.
\newblock In {\em Fifth International Joint Conference on Artificial
  Intelligence.} (1977).

\bibitem{Games:nadolski+2008}
Nadolski, R.~J., Hummel, H.~G., Van Den~Brink, H.~J., Hoefakker, R.~E.,
  Slootmaker, A., Kurvers, H.~J., and Storm, J.
\newblock Emergo: A methodology and toolkit for developing serious games in
  higher education.
\newblock {\em Simulation \& Gaming 39}, 3 (2008), 338--352.

\bibitem{SG:Nau+2003}
Nau, D., Au, T., Ilghami, O., Kuter, U., Murdock, J., Wu, D., and Yaman, F.
\newblock {SHOP2: An HTN Planning System}.
\newblock {\em Journal of Artificial Intelligence Research (JAIR) 20\/} (2003),
  379--404.

\bibitem{SG:NiehausLiRiedl_2010}
Niehaus, J., Li, B., and Riedl, M.
\newblock Automated scenario adaptation in support of intelligent tutoring
  systems.
\newblock In {\em Twenty-Fourth International Florida Artificial Intelligence
  Research Society Conference} (2010).

\bibitem{crowd:AmazonTurk+2010}
Paolacci, G., Chandler, J., and Ipeirotis, P.
\newblock Running experiments on amazon mechanical turk.
\newblock {\em Judgment and Decision Making\/} (2010).

\bibitem{SG:RiedlStern_2006}
Riedl, M., and Stern, A.
\newblock Believable agents and intelligent story adaptation for interactive
  storytelling.
\newblock In {\em Technologies for Interactive Digital Storytelling and
  Entertainment}, Springer (2006), 1--12.

\bibitem{SG:RiedlYoung_2010}
Riedl, M., and Young, R.
\newblock Narrative planning: Balancing plot and character.
\newblock {\em Journal of Artificial Intelligence Research (JAIR) 39\/} (2010),
  217--268.

\bibitem{Games:susi+2007}
Susi, T., Johannesson, M., and Backlund, P.
\newblock Serious games: An overview.

\bibitem{SG:Turner_1993}
Turner, S.
\newblock {MINSTREL: a computer model of creativity and storytelling.}

\bibitem{SG:WareYoung_2011}
Ware, S., and Young, R.
\newblock {CPOCL: A Narrative Planner Supporting Conflict}.
\newblock In {\em Seventh AAAI Conference on Artificial Intelligence and
  Interactive Digital Entertainment, AIIDE} (2011).

\bibitem{SG:Weyhrauch_1997}
Weyhrauch, P., and Bates, J.
\newblock {\em Guiding interactive drama}.
\newblock Carnegie Mellon University, 1997.

\bibitem{SG:ZookRiedl+2012}
Zook, A., Lee-Urban, S., Riedl, M.~O., Holden, H.~K., Sottilare, R.~A., and
  Brawner, K.~W.
\newblock Automated scenario generation: Toward tailored and optimized military
  training in virtual environments.
\newblock In {\em International Conf on the Foundations of Digital Games}, ACM
  (2012).

\end{thebibliography}

\end{document}